\pdfoutput=1
\documentclass[letterpaper, 11pt]{article}
\usepackage[margin = 1in]{geometry}

\usepackage{natbib}
\usepackage{appendix}

\usepackage{amsmath}
\usepackage{amssymb}
\usepackage{amsthm}
\usepackage{booktabs}
\usepackage{xspace}
\usepackage{graphicx}

\usepackage{float}
\usepackage{subfig}

\usepackage{mathtools}
\usepackage{pifont}
\usepackage{cancel}
\usepackage{multirow}
\usepackage{colortbl}
\usepackage{utfsym}

\def\bfa{\mathbf{a}}
\def\bfb{\mathbf{b}}
\def\bfc{\mathbf{c}}

\def\bfr{\mathbf{r}}

\def\bfu{\mathbf{u}}

\def\bfw{\mathbf{w}}
\def\bfx{\mathbf{x}}
\def\bfy{\mathbf{y}}

\def\bfB{\mathbf{B}}

\def\bfI{\mathbf{I}}

\def\bfP{\mathbf{P}}
\def\bfQ{\mathbf{Q}}
\def\bfR{\mathbf{R}}

\def\bfU{\mathbf{U}}

\def\bfW{\mathbf{W}}

\def\rmd{\mathrm{d}}

\def\bbE{\mathbb{E}}
\def\bbR{\mathbb{R}}

\def\nablaxt{\nabla_{\bfx_t}}

\def\calN{\mathcal{N}}
\def\calO{\mathcal{O}}

\def\barx{\bar{\bfx}}
\def\bary{\bar{\bfy}}

\def\bfmu{\boldsymbol{\mu}}

\def\bfI{\mathbf{I}}

\def\eqref#1{(\ref{#1})}

\def\bfSigma{\boldsymbol{\Sigma}}
\def\bfLambda{\boldsymbol{\Lambda}}

\usepackage{hyperref}
\hypersetup{
	colorlinks=true,
	breaklinks=true,
	urlcolor=blue,
	linkcolor=blue,
	bookmarksopen=false,
	filecolor=black,
	citecolor=blue,
	linkbordercolor=blue
}

\graphicspath{{Fig/}}

\theoremstyle{plain}%
\newtheorem{theorem}{Theorem}
\newtheorem{proposition}[theorem]{Proposition}%
\theoremstyle{remark}%
\newtheorem{remark}{Remark}%
\theoremstyle{definition}%

\begin{document}

\title{On the universal structure in deterministic generative diffusion}

\author{Defang Chen$^{1}$ \quad Zhenyu Zhou$^{2}$ \quad Can Wang$^{2}$ \quad Zeyu Zheng$^{1}$ \quad Siwei Lyu$^{3}$
\\$^{1}$UC Berkeley \quad $^{2}$Zhejiang University \quad $^{3}$University at Buffalo
}
\date{\href{mailto:defchern@berkeley.edu,zyzheng@berkeley.edu}{\{\texttt{defchern},\texttt{zyzheng}\}@berkeley.edu} \quad \href{mailto:zhyzhou@zju.edu.cn}{\texttt{zhyzhou}@zju.edu.cn}}

\maketitle

\begingroup
\renewcommand\thefootnote{}        
\footnotetext{An early version of this manuscript (April 2026) is available \href{https://defangchen.github.io/}{here}; the present version was reorganized with the help of Claude Opus 5.5. Code is available \href{https://github.com/zju-pi/GaussianFlows}{here}.
}  
\endgroup

\begin{abstract}
    Diffusion-based generative modeling has emerged as a general computational framework for simulating high-dimensional, structured data across various domains. Understanding its generative mechanisms is paramount for advancing model training and inference in a principled way. Here, we demonstrate that deterministic sampling trajectories share a universal geometric structure invariant to data modalities and diffusion model implementations. This structure can be largely explained by an exactly solvable model based on the optimal linear denoiser.
\end{abstract}

\section{Introduction}

A recurring theme in scientific inquiry is that systems of immense complexity often exhibit unexpectedly simple regularities. 
For example, Kepler's three laws first reduced intricate planetary trajectories to simple geometric form, and Newtonian mechanics then unified celestial and terrestrial dynamics under the same laws of motion. Identifying such universal principles, which transcend the details of individual instances, deepens our understanding and often yields predictive or practical power. Today, this paradigm extends beyond the natural world to artificial neural systems, as exemplified by \textit{double descent}~\citep{belkin2019reconciling} and \textit{scaling laws}~\citep{hestness2017deep,bahri2024explaining}. 

Modern generative diffusion models, rooted in nonequilibrium thermodynamics~\citep{jarzynski1997equilibrium,sohl2015deep}, are among the most successful AI systems to date~\citep{song2021sde}. Their power lies in overcoming a long-standing obstacle in generative modeling, namely the difficulty of representing and sampling from distributions that are at once high-dimensional and richly structured. Instead of modeling the data distribution directly, diffusion models amortize computation over iterative denoising steps that begin from an uninformative Gaussian prior. The denoising direction that guides the data-generating process is learned from the predefined diffusion process that perturbs data into noise. This framework now extends from realistic multimedia synthesis to pressing scientific challenges.

\begin{figure*}[t]
\centering
\includegraphics[width=0.9\linewidth]{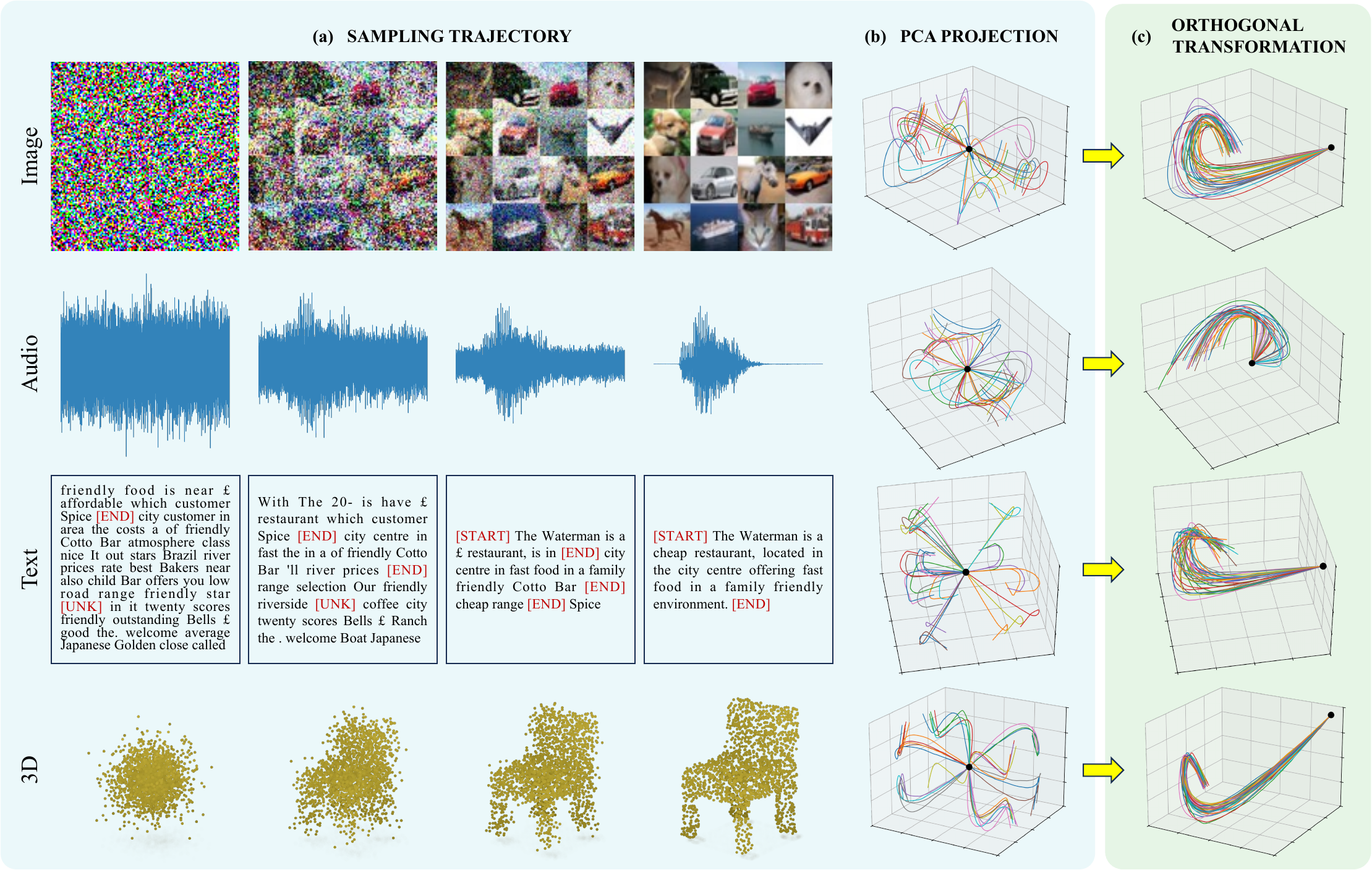}
\caption{Deterministic sampling trajectories share a universal structure across data modalities and diffusion model implementations. (a) Intermediate states along sampling trajectories, from random noise (left) to generated data (right). (b) Each trajectory projected onto its three leading principal components, with all starting points aligned (black point). (c) The trajectories aligned through an orthogonal transformation.  Each axis in (b) and (c) is scaled independently to make the small deviation from a straight line visible.
}
\label{fig:vis}
\end{figure*}

Considering the tremendous complexity of diffusion models and the heterogeneity of the data on which they are trained, it is surprising to us that their \textit{deterministic sampling trajectories follow a common geometric pattern}. Our previous studies identified such a strong regularity in the image domain, covering both unconditional and conditional generation in data and latent spaces~\citep{chen2024trajectory,chen2025geometric}. Whether this pattern reflects the statistics of natural images or is a more general feature of generative diffusion has remained unclear. Here, we confirm its universality through experiments spanning a diverse collection of data modalities and diffusion model implementations (Figure~\ref{fig:vis}). We further show that a simple hyperbolic structure accounts for the dominant trajectory behaviors observed in practice (Figure~\ref{fig:hyperbolic}). This geometric structure naturally arises in an exactly solvable model built on the linear minimum mean-square-error (LMMSE) denoiser, which coincides at every noise level with the Bayes-optimal denoiser for a Gaussian with the same mean and covariance as the data. We call this model the Gaussian flow and discuss its asymptotic trajectory congruence in high dimensions. The Gaussian flow has previously been adopted to connect diffusion models with optimal transport~\citep{khrulkov2023understanding}, to derive Wasserstein errors of numerical samplers~\citep{pierret2025diffusion}, to decompose the denoising direction for low-dimensional data~\citep{chen2023approximation}, and to characterize the trajectory deviation from a straight line in our previous work~\citep{chen2025geometric}. However, none of them considers its hyperbolic structure or the congruence of its trajectories.

\section{Results}

We analyze sampling trajectories across fundamentally different data modalities, namely images~\citep{karras2022edm}, audio~\citep{kong2021diffwave}, text~\citep{li2022diffusionlm}, and 3D point clouds~\citep{zhou20213d}, each with a representative implementation. Details are provided in the Supplementary material. Figure~\ref{fig:vis}(a) visualizes several intermediate states of the deterministic sampling trajectories in data spaces with thousands of dimensions, where various data structures emerge gradually from random noise. We then project each randomly initialized trajectory onto its three leading principal components (Figure~\ref{fig:vis}(b)). The first component alone captures more than 99\% of the trajectory variance for all four models, i.e., each trajectory is nearly a straight line from noise toward data. 
However, a straight line cannot describe how the data structure emerges along the trajectory; this information lies in the small deviation from the line, which traces a characteristic curve in the second and third components. We further align each trajectory with an arbitrarily chosen reference trajectory through an orthogonal transformation. As shown in Figure~\ref{fig:vis}(c), the aligned trajectories share approximately the same geometric shape, which persists across heterogeneous data modalities and model implementations. Such persistence makes it unlikely that these regularities originate in the statistics of any particular dataset or in the design of any particular model implementation. They point instead to a general mechanism intrinsic to how generative diffusion transports distributions. We next take a step toward understanding this universal structure by investigating an exactly solvable model that yields closed-form solutions for deterministic sampling trajectories. 

\subsection{Gaussian reference frame}
Without loss of generality~\citep{chen2025geometric}, we consider the simplest forward diffusion process $\rmd \bfx_t = \sqrt{2t}\,\rmd\bfw_t$. Here, $\bfx_0\sim p_{\text{data}}$ is a data sample from a centered data distribution in $\bbR^d$, and $\bfw_t$ is a standard Wiener process. The sample is corrupted by isotropic Gaussian noise of standard deviation $t$, i.e., $\bfx_t\mid\bfx_0\sim\calN(\bfx_0,t^2\bfI_d)$.
 At a sufficiently large time $T$, the noise dominates the data structure, and $\bfx_T$ approximately follows $\calN(\mathbf{0},T^2\bfI_d)$. The (real) sampling trajectory is generated by solving the deterministic probability flow $\rmd \bfx_t=(1/t)(\bfx_t-\bbE(\bfx_0\mid\bfx_t))\, \rmd t$ backward, where the conditional expectation is estimated using a deep neural network (the denoiser) trained on a large-scale dataset. If we instead estimate $\bbE(\bfx_0\mid\bfx_t)$ with the LMMSE denoiser, or equivalently model the data as Gaussian, we have a deterministic Gaussian flow $\rmd \barx_t =t(\bfSigma+t^2\bfI_d)^{-1}\,\barx_t \,\rmd t$. Here, $\bfSigma$ denotes the data covariance matrix and admits the spectral decomposition $\bfSigma= \bfU\bfLambda \bfU^{\top}$, where the diagonal matrix $\bfLambda=\operatorname{diag}(\lambda_1,\ldots,\lambda_d)$ contains the non-negative eigenvalues sorted in descending order, and the orthogonal matrix $\bfU = [\bfu_1,\ldots,\bfu_d] \in \bbR^{d\times d}$ contains the
corresponding eigenvectors.

\begin{figure*}[t]
    \centering
    \subfloat[Image. 
    \label{fig:hyperbolic_cifar10}]{%
        \begin{minipage}[b]{0.235\textwidth}%
            \includegraphics[width=\linewidth]{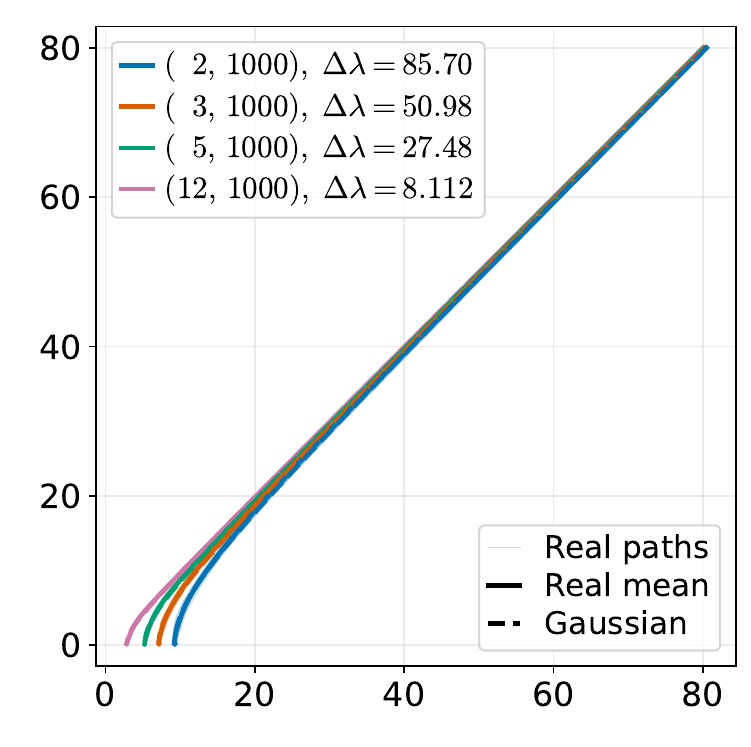}\\[-6pt]%
            \includegraphics[width=\linewidth]{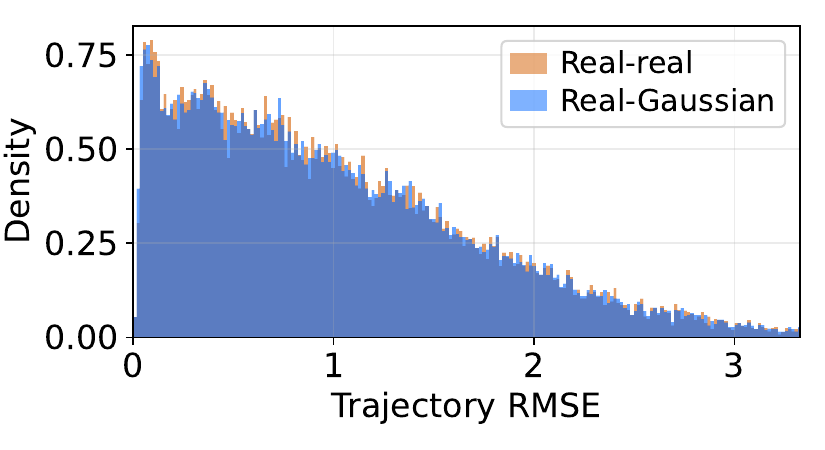}%
        \end{minipage}%
    }%
    \hfill
    \subfloat[Audio. 
    \label{fig:hyperbolic_diffwave}]{%
        \begin{minipage}[b]{0.235\textwidth}%
            \includegraphics[width=\linewidth]{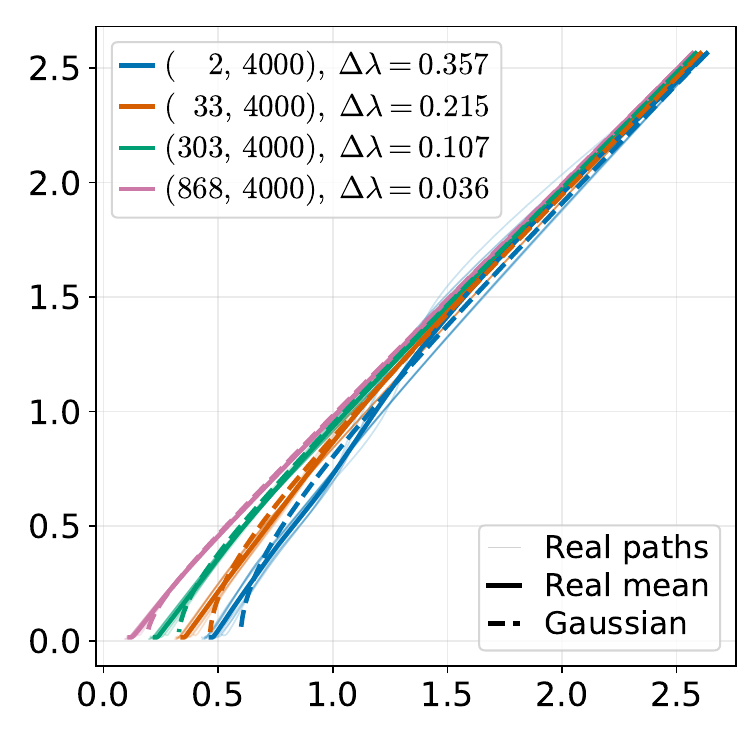}\\[-6pt]%
            \includegraphics[width=\linewidth]{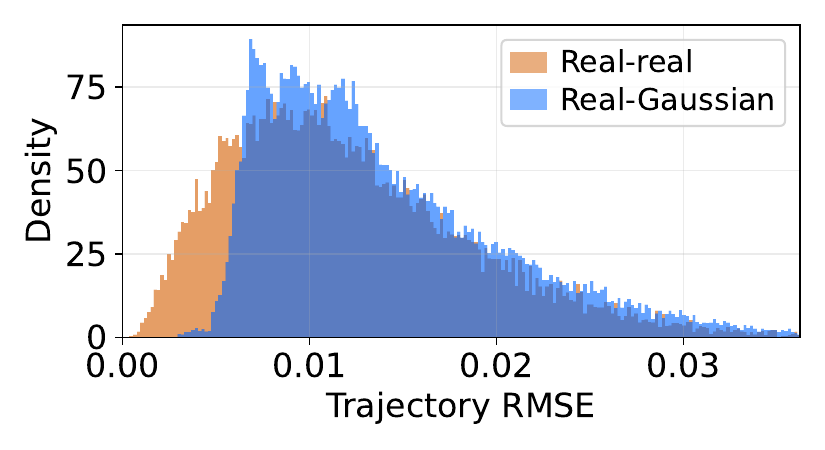}%
        \end{minipage}%
    }%
    \hfill
    \subfloat[Text. 
    \label{fig:hyperbolic_diffusion_lm}]{%
        \begin{minipage}[b]{0.235\textwidth}%
            \includegraphics[width=\linewidth]{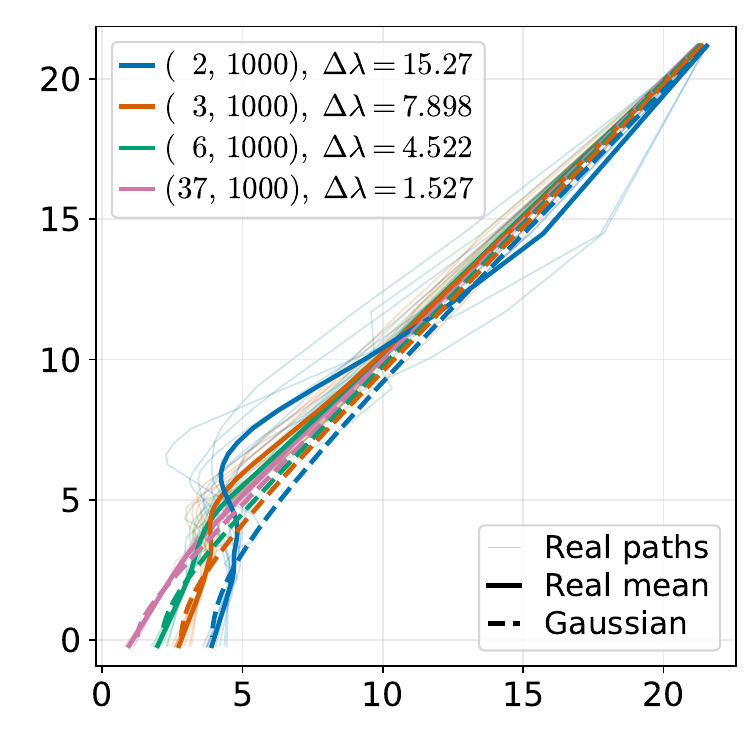}\\[-6pt]%
            \includegraphics[width=\linewidth]{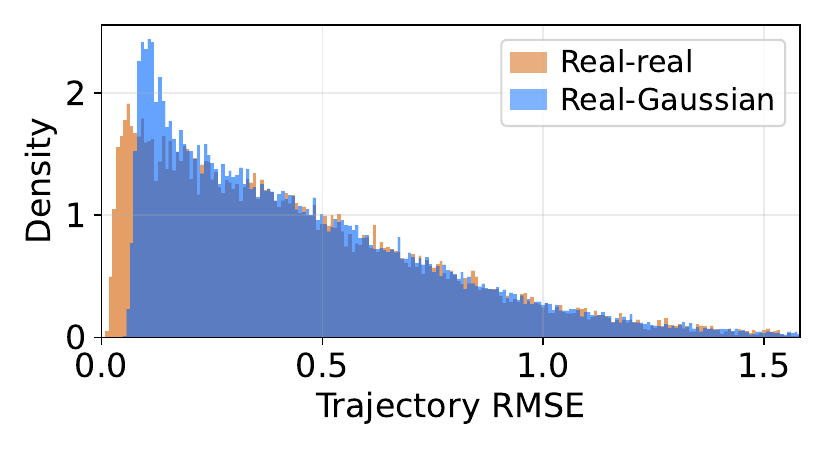}%
        \end{minipage}%
    }%
    \hfill
    \subfloat[3D.
    \label{fig:hyperbolic_pvd}]{%
        \begin{minipage}[b]{0.235\textwidth}%
            \includegraphics[width=\linewidth]{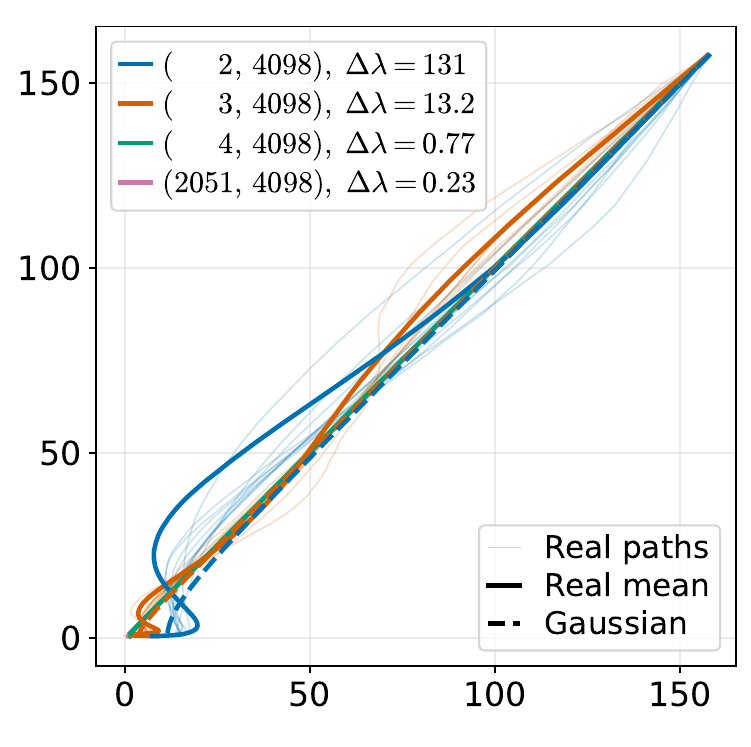}\\[-6pt]%
            \includegraphics[width=\linewidth]{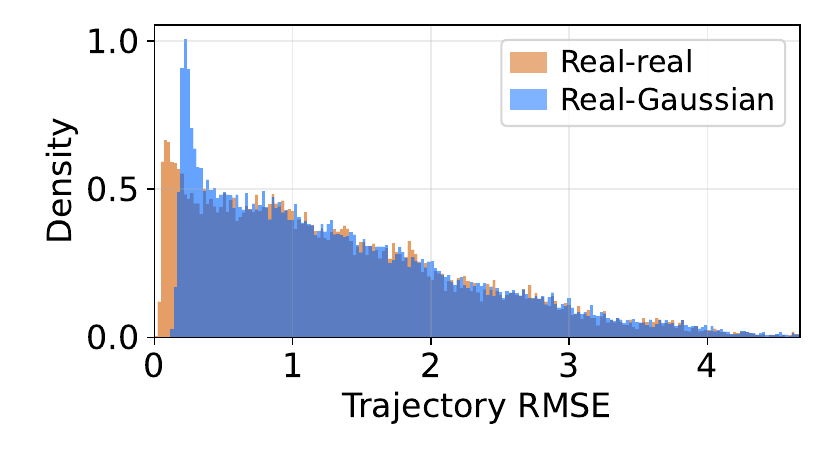}%
        \end{minipage}%
    }%

    \vspace{0.3em}
    {
    \footnotesize
    \begin{tabular}{l*{8}{w{c}{1.25cm}}}
        \toprule
        & \multicolumn{2}{c}{Gaussian chord} & \multicolumn{2}{c}{Brownian bridge} & \multicolumn{2}{c}{Isotropic Gaussian flow} & \multicolumn{2}{c}{Gaussian flow} \\
        \cmidrule(lr){2-3}\cmidrule(lr){4-5}\cmidrule(lr){6-7}\cmidrule(lr){8-9}
        & Before & After & Before & After & Before & After & Before & After \\
        \midrule
        Image~\citep{karras2022edm} & 1 & 0.6157 & 1.4136 & 0.7535 & 1.5703 & 0.9960 & 0.5600 & \textbf{0.1148} \\
        Audio~\citep{kong2021diffwave} & 1 & 0.3736 & 1.4147 & 0.7479 & 0.5031 & 0.3039 & 0.5731 & \textbf{0.1581} \\
        Text~\citep{li2022diffusionlm} & 1 & 0.6241 & 1.4148 & 0.7610 & 0.7669 & 0.2941 & 0.7122 & \textbf{0.1780} \\
        3D~\citep{zhou20213d} & 1 & 0.9047 & 1.4138 & 1.0307 & 0.7361 & 0.4189 & 0.5873 & \textbf{0.3419} \\
        \bottomrule
    \end{tabular}
    }
    \caption{The real trajectories, after an orthogonal transformation, closely follow the Gaussian hyperbolas for most eigenmode pairs (top row; legend: $(i,j)$ and the eigenvalue gap $\Delta\lambda=\lambda_i-\lambda_j$), except at the largest $\Delta\lambda$. The bottom row shows the distributions of the shape deviation, measured by the root-mean-square error (RMSE), between two independently initialized real trajectories and between a real trajectory and the Gaussian counterpart of another such trajectory.
    The table reports the deviation of each reference from the real trajectory before and after the transformation, normalized by the Gaussian chord deviation before the transformation. The Gaussian chord is the straight line from the initial noise to the endpoint of the Gaussian flow, the Brownian bridge is pinned to the same two endpoints, and the isotropic Gaussian flow replaces $\bfSigma$ with the scaled identity $(\operatorname{tr}\bfSigma/d)\,\bfI_d$ (see the Supplementary material for implementation details).} 
    \label{fig:hyperbolic}
\end{figure*}

One interesting property of such an exactly solvable model is that its induced Gaussian sampling trajectory has an invariant hyperbolic structure irrespective of the initial noise. In particular, the solution in the spectral representation ($\bary_{t} = \bfU^{\top}\barx_{t}$) evolves independently across dimensions and decouples the deterministic and stochastic components: the $i$-th coordinate $\bar{y}_{t,i}=\alpha_{t,i}\,\tilde{y}_{T,i}$ tracks the restoration of data features with a deterministic amplitude $\alpha_{t,i}\coloneq \sqrt{\lambda_i+t^2}$ across different realizations, and the initial stochasticity $\tilde{y}_{T,i}$, which approximately follows $\calN(0, 1)$, acts as a scaling factor for each coordinate. 
Furthermore, projecting the trajectory onto the plane spanned by $\bfu_i$ and $\bfu_j$ with non-zero initial conditions ($\tilde{y}_{T,i}\, \tilde{y}_{T,j} \neq 0$) yields a~hyperbolic~segment
\begin{align}
    \label{eq:hyperbola}
    (\bar{y}_{t,i}/\tilde{y}_{T,i})^2-(\bar{y}_{t,j}/\tilde{y}_{T,j})^2=
    \alpha_{t,i}^2 - \alpha_{t,j}^2&=\lambda_i-\lambda_j,
\end{align}
with $t\in [0, T]$. Each eigenvalue gap thus defines one hyperbola, and these hyperbolas form a nested family that shares the diagonal as its asymptote. Geometrically, each relation (Eq.~\ref{eq:hyperbola}) defines a hyperbolic cylinder in $\bbR^{d}$, and the intersection of these $(d-1)$ hyperbolic cylinders (if non-degenerate) confines the generative dynamics to a one-dimensional trajectory $\bary_t=(\bfLambda+t^2\bfI_d)^{1/2}\tilde{\bfy}_T$. The trajectory's curvature vanishes as $t \to \infty$, and, with $t$ uniformly distributed on $[0,T]$, its leading principal component already captures all but a fraction $\calO(T^{-3})$ of its variance (see the Supplementary material). This invariant structure makes the Gaussian flow suitable as a yardstick to measure the trajectory variation and the nonlinear deviation.
 
In this Gaussian reference frame, each real trajectory closely follows these hyperbolas when the eigenvalue gap is moderate (Figure~\ref{fig:hyperbolic}, top row). This comparison, like the ``After'' columns of the table, examines the trajectory shape rather than its evolution along the original directions (``Before'' columns). After the transformation, each real trajectory deviates on average less from its Gaussian counterpart than from the other three references, and about as much from another Gaussian trajectory as from another real one (Figure~\ref{fig:hyperbolic}, bottom row). 

Furthermore, Gaussian sampling trajectories starting from distinct initial states are approximately congruent (equivalent under orthogonal transformations) in high dimensions. The \textit{relative fluctuation} of the Gram kernel $G(t,s)=\bary_t^{\top}\bary_{s}$ quantifies how far a single realization deviates from its expectation over the initial noise:
\begin{align}
    \label{eq:fluctuation}
    \frac{\sqrt{\operatorname{Var}\left(G(t,s)\right)}}{\mathbb{E}\left[G(t,s)\right]} = \frac{\sqrt{2\sum_{i=1}^{d}(\lambda_i+t^2)(\lambda_i+s^2)}}{\sum_{i=1}^d \sqrt{(\lambda_i+t^2)(\lambda_i+s^2)}}.
\end{align}
The fluctuation depends on the data eigenspectrum and the noise levels. Along the diagonal $t=s$, it gradually transitions from $\mathcal{O}(d^{-1/2})$ to $\mathcal{O}(r_{\text{eff}}^{-1/2})$ as the reverse process travels from pure noise to the data end. Here, $r_{\text{eff}}\coloneq (\operatorname{tr}\bfLambda)^2/\operatorname{tr}(\bfLambda^2)$ is the effective data dimension. For a power-law eigenspectrum $\lambda_i\propto i^{-\alpha}$, as is typical of natural images, $r_{\text{eff}}^{-1/2}$ vanishes as $d\to\infty$ when $\alpha\le1$ and otherwise tends to a positive constant (see the Supplementary material). Although the fluctuation increases near the data end, rendering the trajectory geometry slightly more varied, the overall structural congruence remains robust.

\section{Conclusion}

Elucidating the geometric structure of generative diffusion dynamics is fundamental to explaining empirical observations and developing principled techniques for model training and inference. We find that deterministic sampling trajectories share approximately the same geometric shape across data modalities and model implementations, beyond the image domain of our previous studies~\citep{chen2024trajectory,chen2025geometric}. Using the Gaussian flow built on the LMMSE denoiser, we rigorously characterize the hyperbolic structure of its sampling trajectories in high dimensions and show that this structure approximately recovers the observed shape. Our findings open new avenues for the design of diffusion-based generative models, including the identification of optimal sampling configurations and the acceleration of model training through prior knowledge of the trajectory structure.

\section{Author contributions}

D.C.: Conceptualization, Methodology, Formal analysis, Writing -- original draft; Z. Zhou: Software, Investigation, Visualization, Writing -- review \& editing; C.W., Z. Zheng, S.L.: Resources, Supervision, Writing -- review \& editing.

\newpage

\bibliographystyle{plainnat}
\bibliography{reference}

\begin{thebibliography}{15}
\providecommand{\natexlab}[1]{#1}
\providecommand{\url}[1]{\texttt{#1}}
\expandafter\ifx\csname urlstyle\endcsname\relax
  \providecommand{\doi}[1]{doi: #1}\else
  \providecommand{\doi}{doi: \begingroup \urlstyle{rm}\Url}\fi

\bibitem[Bahri et~al.(2024)Bahri, Dyer, Kaplan, Lee, and
  Sharma]{bahri2024explaining}
Yasaman Bahri, Ethan Dyer, Jared Kaplan, Jaehoon Lee, and Utkarsh Sharma.
\newblock Explaining neural scaling laws.
\newblock \emph{Proceedings of the National Academy of Sciences}, 121\penalty0
  (27):\penalty0 e2311878121, 2024.

\bibitem[Belkin et~al.(2019)Belkin, Hsu, Ma, and Mandal]{belkin2019reconciling}
Mikhail Belkin, Daniel Hsu, Siyuan Ma, and Soumik Mandal.
\newblock Reconciling modern machine-learning practice and the classical
  bias--variance trade-off.
\newblock \emph{Proceedings of the National Academy of Sciences}, 116\penalty0
  (32):\penalty0 15849--15854, 2019.

\bibitem[Chen et~al.(2024)Chen, Zhou, Wang, Shen, and Lyu]{chen2024trajectory}
Defang Chen, Zhenyu Zhou, Can Wang, Chunhua Shen, and Siwei Lyu.
\newblock On the trajectory regularity of {ODE}-based diffusion sampling.
\newblock In \emph{International Conference on Machine Learning}, volume 235,
  pages 7905--7934. PMLR, 2024.

\bibitem[Chen et~al.(2025)Chen, Zhou, Wang, and Lyu]{chen2025geometric}
Defang Chen, Zhenyu Zhou, Can Wang, and Siwei Lyu.
\newblock Geometric regularity in deterministic sampling dynamics of
  diffusion-based generative models.
\newblock \emph{Journal of Statistical Mechanics: Theory and Experiment},
  2025\penalty0 (12):\penalty0 124002, 2025.

\bibitem[Chen et~al.(2023)Chen, Huang, Zhao, and Wang]{chen2023approximation}
Minshuo Chen, Kaixuan Huang, Tuo Zhao, and Mengdi Wang.
\newblock Score approximation, estimation and distribution recovery of
  diffusion models on low-dimensional data.
\newblock In \emph{International Conference on Machine Learning}, volume 202,
  pages 4672--4712. PMLR, 2023.

\bibitem[Hestness et~al.(2017)Hestness, Narang, Ardalani, Diamos, Jun,
  Kianinejad, Patwary, Yang, and Zhou]{hestness2017deep}
Joel Hestness, Sharan Narang, Newsha Ardalani, Gregory Diamos, Heewoo Jun,
  Hassan Kianinejad, Md~Mostofa~Ali Patwary, Yang Yang, and Yanqi Zhou.
\newblock Deep learning scaling is predictable, empirically.
\newblock \emph{arXiv preprint arXiv:1712.00409}, 2017.

\bibitem[Jarzynski(1997)]{jarzynski1997equilibrium}
Christopher Jarzynski.
\newblock Equilibrium free-energy differences from nonequilibrium measurements:
  A master-equation approach.
\newblock \emph{Physical Review E}, 56\penalty0 (5):\penalty0 5018, 1997.

\bibitem[Karras et~al.(2022)Karras, Aittala, Aila, and Laine]{karras2022edm}
Tero Karras, Miika Aittala, Timo Aila, and Samuli Laine.
\newblock Elucidating the design space of diffusion-based generative models.
\newblock In \emph{Advances in Neural Information Processing Systems},
  volume~35, pages 26565--26577, 2022.

\bibitem[Khrulkov et~al.(2023)Khrulkov, Ryzhakov, Chertkov, and
  Oseledets]{khrulkov2023understanding}
Valentin Khrulkov, Gleb Ryzhakov, Andrei Chertkov, and Ivan Oseledets.
\newblock Understanding {DDPM} latent codes through optimal transport.
\newblock In \emph{International Conference on Learning Representations}, 2023.

\bibitem[Kong et~al.(2021)Kong, Ping, Huang, Zhao, and
  Catanzaro]{kong2021diffwave}
Zhifeng Kong, Wei Ping, Jiaji Huang, Kexin Zhao, and Bryan Catanzaro.
\newblock {DiffWave}: {A} versatile diffusion model for audio synthesis.
\newblock In \emph{International Conference on Learning Representations}, 2021.

\bibitem[Li et~al.(2022)Li, Thickstun, Gulrajani, Liang, and
  Hashimoto]{li2022diffusionlm}
Xiang Li, John Thickstun, Ishaan Gulrajani, Percy~S Liang, and Tatsunori~B
  Hashimoto.
\newblock {Diffusion-LM} improves controllable text generation.
\newblock In \emph{Advances in Neural Information Processing Systems},
  volume~35, pages 4328--4343, 2022.

\bibitem[Pierret and Galerne(2025)]{pierret2025diffusion}
Emile Pierret and Bruno Galerne.
\newblock Diffusion models for {G}aussian distributions: Exact solutions and
  {W}asserstein errors.
\newblock In \emph{International Conference on Machine Learning}, volume 267,
  pages 49355--49381, 2025.

\bibitem[Sohl-Dickstein et~al.(2015)Sohl-Dickstein, Weiss, Maheswaranathan, and
  Ganguli]{sohl2015deep}
Jascha Sohl-Dickstein, Eric Weiss, Niru Maheswaranathan, and Surya Ganguli.
\newblock Deep unsupervised learning using nonequilibrium thermodynamics.
\newblock In \emph{International Conference on Machine Learning}, volume~37,
  pages 2256--2265. PMLR, 2015.

\bibitem[Song et~al.(2021)Song, Sohl{-}Dickstein, Kingma, Kumar, Ermon, and
  Poole]{song2021sde}
Yang Song, Jascha Sohl{-}Dickstein, Diederik~P. Kingma, Abhishek Kumar, Stefano
  Ermon, and Ben Poole.
\newblock Score-based generative modeling through stochastic differential
  equations.
\newblock In \emph{International Conference on Learning Representations}, 2021.

\bibitem[Zhou et~al.(2021)Zhou, Du, and Wu]{zhou20213d}
Linqi Zhou, Yilun Du, and Jiajun Wu.
\newblock {3D} shape generation and completion through point-voxel diffusion.
\newblock In \emph{Proceedings of the IEEE/CVF International Conference on
  Computer Vision}, pages 5826--5835, 2021.

\end{thebibliography}

\begin{appendices}

\section{Supplementary material}

\subsection{Generative diffusion dynamics} 

We consider a diffusion process $\{\bfx_{t}\}_{t\in [0,T]}$ in $\bbR^d$, governed by the linear stochastic differential equation $\rmd \bfx_t = f(t)\bfx_t\rmd t + g(t)\, \rmd \bfw_t$, where $\bfw_{t}$ is the Wiener process, and $f(t)$ and $g(t)$ are the drift and diffusion coefficients, respectively. The initial condition $\bfx_0$ is drawn from a data distribution $p_0$, which can be, for example, a distribution of images, audio, or text. The marginal density $p_t$ evolves according to the Fokker--Planck equation $\partial p_t(\bfx_t)/ \partial t = - \nabla \cdot [f(t)\bfx_tp_t(\bfx_t)  -(g^2(t)/2)\nablaxt p_t(\bfx_t)]$. As $t\rightarrow T$, the data distribution is smoothly transformed into an approximately Gaussian distribution for suitably chosen $f(t)$ and $g(t)$. This stochastic diffusion process can be reversed with the (Stein) score function $\nablaxt \log p_t(\bfx_t)$, i.e., the gradient of the log marginal density with respect to $\bfx_t$. 
The property that makes fast data generation possible is that the same marginal evolution can be reproduced by a \textit{deterministic} process characterized by the probability flow ordinary differential equation, 
\begin{equation}
    \label{eq:pf_ode}
    \rmd \bfx_t
		= \left[f(t)\bfx_t - \frac{1}{2} g^2(t) \nablaxt \log p_t(\bfx_t)\right] \rmd t.
\end{equation}
Integrating Eq.~\ref{eq:pf_ode} backward in time from initial Gaussian noise yields the synthesized datum and the deterministic sampling trajectory, whose geometry is the central object of this paper.

This formulation underlies most diffusion- and flow-based models, which use various designs of the drift and diffusion coefficients, as well as iterative samplers based on numerical solvers that discretize Eq.~\ref{eq:pf_ode} into reliable finite-step sampling trajectories. In practice, the score function is estimated using a deep neural network trained on a large-scale dataset with denoising score matching or equivalent objectives~\citep{song2021sde}. More details are provided in our previous study~\citep{chen2025geometric}. 

\subsection{Gaussian flows}

We collect the notation of the deterministic Gaussian flow used in the main text. The data have mean $\bfmu=\mathbf{0}$ and covariance matrix $\bfSigma\in\bbR^{d\times d}$, which is positive semi-definite with the spectral decomposition $\bfSigma=\bfU\bfLambda\bfU^{\top}$, where $\bfLambda=\operatorname{diag}(\lambda_1,\ldots,\lambda_d)$ with $\lambda_1\ge\cdots\ge\lambda_d\ge 0$, and $\bfU=[\bfu_1,\ldots,\bfu_d]\in\bbR^{d\times d}$ is orthogonal. Starting from an initial state $\barx_T$, the deterministic Gaussian flow $\rmd\barx_t=t(\bfSigma+t^2\bfI_d)^{-1}\barx_t\,\rmd t$ has the closed-form solution
\begin{align}
    \label{eq:flow_map}
    \barx_t=\left(\bfSigma+t^2\bfI_d\right)^{1/2}\left(\bfSigma+T^2\bfI_d\right)^{-1/2}\barx_T,\quad t\in[0,T].
\end{align}
For data with a nonzero mean $\bfmu$, as in our experiments, Eq.~\ref{eq:flow_map} applies to $\barx_t-\bfmu$ and $\barx_T-\bfmu$.
In the eigenbasis, $\bary_t=\bfU^{\top}\barx_t$ evolves coordinate-wise as $\bar{y}_{t,i}=\alpha_{t,i}\, \tilde{y}_{T,i}$, where $\alpha_{t,i}\coloneq\sqrt{\lambda_i+t^2}$ and $\tilde{y}_{T,i}\coloneq\bar{y}_{T,i}/\alpha_{T,i}$. When $\barx_T\sim\calN(\mathbf{0},\bfSigma+T^2\bfI_d)$\footnote{In practice, we initialize the reverse process from the isotropic noise distribution $\calN(\mathbf{0}, T^2\bfI_d)$, which accurately approximates $p_T$ provided $T$ is sufficiently large to dominate the data structure, and this determinism-stochasticity dichotomy still holds.}, we have $\tilde{\bfy}_T\sim\calN(\mathbf{0}, \bfI_d)$. The trajectory deviates only slightly from a straight line, since all coordinates share the common linear growth, as captured by the Taylor expansion $\bary_{t} \approx (t\bfI_d + \bfLambda/(2t)) {\tilde{\bfy}_T}$ for $t\to \infty$. 


\subsubsection{Straight-line decomposition} 
We split the deterministic amplitude $\alpha_{t,i}$ into a common linear part and a bounded correction, $\alpha_{t,i}=t+\rho_{t,i}$, with
\begin{align}
    \label{eq:rho}
    \rho_{t,i}\coloneq\sqrt{\lambda_i+t^2}-t=\frac{\lambda_i}{\sqrt{\lambda_i+t^2}+t}.
\end{align}
The correction $\rho_{t,i}$ has three properties: (i) it decreases in $t$ from its maximum $\rho_{0,i}=\sqrt{\lambda_i}$ at the data end, since $\partial\rho_{t,i}/\partial t=t/\sqrt{\lambda_i+t^2}-1\le0$, and thus $\rho_{t,i}\in[0,\sqrt{\lambda_i}\,]$; (ii) $\rho_{t,i}\le\lambda_i/(2t)$ and $\rho_{t,i}=\lambda_i/(2t)+\calO(t^{-3})$ for $t\gg\sqrt{\lambda_i}$; (iii) it increases in $\lambda_i$, since $\partial\rho_{t,i}/\partial\lambda_i=1/(2\sqrt{\lambda_i+t^2})>0$, and vanishes when $\lambda_i=0$.

The same split holds for the matrix square root in Eq.~\ref{eq:flow_map}, i.e., $\left(\bfSigma+t^2\bfI_d\right)^{1/2}=t\,\bfI_d+\bfR_t$, where $\bfR_t\coloneq\bfU\operatorname{diag}\big(\rho_{t,1},\ldots,\rho_{t,d}\big)\bfU^{\top}$ is positive semi-definite and shares the eigenvectors of $\bfSigma$. Substituting this into Eq.~\ref{eq:flow_map} gives
\begin{align}
    \barx_t
    =\left(t\,\bfI_d+\bfR_t\right)\left(\bfSigma+T^2\bfI_d\right)^{-1/2}\barx_T=t\,\bfa+\bfr_t,
    \label{eq:line_decomp}
\end{align}
where
\begin{align}
    \bfa\coloneq\left(\bfSigma+T^2\bfI_d\right)^{-1/2}\barx_T=\bfU\tilde{\bfy}_T, \quad \bfr_t\coloneq\bfR_t\bfa.
\end{align}
The straight part $t\bfa$ thus depends only on the initial noise and vanishes at $t=0$, i.e., the line passes through the data mean (the origin), so the generated structure is carried entirely by the deviation $\bfr_t$, to which directions with $\lambda_i=0$ do not contribute (property (iii)). Away from the data end, this deviation is small relative to the straight part: by property (iii), $\bfR_t$ has the largest eigenvalue $\rho_{t,1}$, so $\|\bfr_t\|_2\le\rho_{t,1}\|\bfa\|_2$, and property (ii) gives
\begin{align}
    \label{eq:rel_dev}
    \frac{\|\bfr_t\|_2}{\|t\,\bfa\|_2}\le\frac{\rho_{t,1}}{t}\le\frac{\lambda_1}{2t^2},
\end{align}
which is below one for $t>\sqrt{\lambda_1/2}$.

\paragraph{Temporal statistics.} Let $\nu$ be a probability measure on $[0,T]$ that weights the time points of the trajectory, e.g., the uniform distribution for the continuous-time trajectory or the empirical distribution of the time steps of a numerical sampler. We write $\bbE_t$, $\operatorname{Var}_t$, and $\operatorname{Cov}_t$ for the mean, variance, and covariance over $t\sim\nu$. The principal components of the trajectory are the eigenvectors of its temporal covariance
\begin{align}
    \bfSigma_{\text{path}}\coloneq\operatorname{Cov}_t(\barx_t)=\bbE_t\left[(\barx_t-\bbE_t\barx_t)(\barx_t-\bbE_t\barx_t)^{\top}\right],
\end{align}
with eigenvalues $\mu_1\ge\cdots\ge\mu_d\ge0$. We measure the fraction of the trajectory variance not captured by its $k$ leading principal components by $\varepsilon_k\coloneq\sum_{j>k}\mu_j/\sum_{j=1}^{d}\mu_j$.

\begin{proposition}
\label{prop:lowrank}
For any positive semi-definite $\bfSigma$, any initial state with $\tilde{\bfy}_T\neq\mathbf{0}$, and any probability measure $\nu$ on $[0,T]$ with $\operatorname{Var}_t(t)>0$, define
\begin{align}
    \delta\coloneq\frac{\bbE_t\|\bfr_t\|_2^2}{\operatorname{Var}_t(t)\,\|\bfa\|_2^2}
    =\frac{\sum_{i=1}^{d}\tilde{y}_{T,i}^2\,\bbE_t\big[\rho_{t,i}^2\big]}{\operatorname{Var}_t(t)\sum_{i=1}^{d}\tilde{y}_{T,i}^2}.
\end{align}
\textit{(a)} If $\delta<1$, then $\varepsilon_2\le\varepsilon_1\le\delta/(1-\sqrt{\delta})^2$. \textit{(b)} If $\nu$ is the uniform distribution on $[0,T]$, then
\begin{align}
    \delta\le 8\,\frac{\sum_{i=1}^{d}\tilde{y}_{T,i}^2\left(\lambda_i/T^2\right)^{3/2}}{\sum_{i=1}^{d}\tilde{y}_{T,i}^2}\le 8\left(\frac{\lambda_1}{T^2}\right)^{3/2}.
\end{align}
Hence, $\varepsilon_2\le\varepsilon_1=\calO(T^{-3})$ as $T\to\infty$.
\end{proposition}

\begin{proof}
\textit{(a)} Let $\hat{\bfa}=\bfa/\|\bfa\|_2$ and $\bfP=\bfI_d-\hat{\bfa}\hat{\bfa}^{\top}$. By the variational characterization of the leading eigenvalue, $\mu_1\ge\hat{\bfa}^{\top}\bfSigma_{\text{path}}\hat{\bfa}$, and thus
\begin{align}
    \sum_{j=2}^{d}\mu_j
    &=\operatorname{tr}(\bfSigma_{\text{path}})-\mu_1
    \le\operatorname{tr}\left(\bfP\bfSigma_{\text{path}}\bfP\right)\nonumber\\
    &=\bbE_t\big\|\bfP(\bfr_t-\bbE_t\bfr_t)\big\|_2^2
    \le\bbE_t\|\bfr_t\|_2^2,
\end{align}
where the equality uses $\bfP\barx_t=\bfP\bfr_t$ (Eq.~\ref{eq:line_decomp}), and the last step uses that $\bfP$ is an orthogonal projection and that $\bbE_t\|\bfr_t-\bbE_t\bfr_t\|_2^2=\bbE_t\|\bfr_t\|_2^2-\|\bbE_t\bfr_t\|_2^2$. For the total variance, the triangle inequality in $L^2(\nu)$ gives
\begin{align}
    \sqrt{\operatorname{tr}(\bfSigma_{\text{path}})}
    &=\Big(\bbE_t\big\|(t-\bbE_t t)\,\bfa+(\bfr_t-\bbE_t\bfr_t)\big\|_2^2\Big)^{1/2}\nonumber\\
    &\ge\sqrt{\operatorname{Var}_t(t)}\,\|\bfa\|_2-\sqrt{\bbE_t\|\bfr_t\|_2^2}\nonumber\\
    &=\left(1-\sqrt{\delta}\right)\sqrt{\operatorname{Var}_t(t)}\,\|\bfa\|_2,
\end{align}
which is positive when $\delta<1$. Dividing the first bound by the square of the second yields $\varepsilon_1\le\delta/(1-\sqrt{\delta})^2$, and $\varepsilon_2\le\varepsilon_1$ holds since $\mu_2\ge0$.

\textit{(b)} For the uniform distribution, $\operatorname{Var}_t(t)=T^2/12$. The function $F(t)=\lambda_i t+\tfrac{2}{3}t^3-\tfrac{2}{3}(\lambda_i+t^2)^{3/2}$ satisfies $F'(t)=\rho_{t,i}^2$, $F(0)=-\tfrac{2}{3}\lambda_i^{3/2}$, and $F(t)\to0$ as $t\to\infty$. Therefore,
\begin{align}
    \bbE_t\big[\rho_{t,i}^2\big]=\frac{1}{T}\int_0^T\rho_{t,i}^2\,\rmd t\le\frac{1}{T}\int_0^\infty\rho_{t,i}^2\,\rmd t=\frac{2\lambda_i^{3/2}}{3T}.
\end{align}
Substituting this bound and $\operatorname{Var}_t(t)=T^2/12$ into $\delta$ gives the first inequality, and $\lambda_i\le\lambda_1$ gives the second. Since $\delta=\calO(T^{-3})$, part (a) yields $\varepsilon_1=\calO(T^{-3})$.
\end{proof}

Proposition~\ref{prop:lowrank} holds for every realization of the initial noise and requires no low-rank assumption on $\bfSigma$. Part (a) also applies to a principal component analysis that weights every saved state equally, in which case $\nu$ is the empirical distribution of their time steps, so that $\delta$, and hence the bound on $\varepsilon_1$, depends on the schedule. A schedule that places most of its steps near the data end, such as that of EDM~\citep{karras2022edm}, gives more weight to small $t$, where $\rho_{t,i}$ is largest (property (i)), and increases the numerator of $\delta$. Weighting each state by its step size instead makes $\nu$ nearly uniform, as in part (b). 

\subsection{Implementation details}

\subsubsection{Models and common protocol}
We consider four data modalities and use a representative unconditional diffusion model for each: images generated by EDM~\citep{karras2022edm} on CIFAR-10\footnote{\url{https://www.cs.toronto.edu/~kriz/cifar.html}}, raw audio waveforms generated by DiffWave~\citep{kong2021diffwave} on SC09\footnote{\url{http://download.tensorflow.org/data/speech_commands_v0.02.tar.gz}}, text generated by Diffusion-LM~\citep{li2022diffusionlm} on E2E\footnote{\url{https://github.com/tuetschek/e2e-dataset}}, and 3D point clouds generated by PVD~\citep{zhou20213d} on the Chair category of ShapeNet\footnote{\url{https://github.com/stevenygd/PointFlow}}. We use the official implementations and their publicly released checkpoints\footnote{\url{https://nvlabs-fi-cdn.nvidia.com/edm/pretrained/edm-cifar10-32x32-uncond-vp.pkl}}\textsuperscript{,}\footnote{\url{https://github.com/albertfgu/diffwave-sashimi/blob/checkpoints/exp/unet_d128_n6_pool_2_expand2_ff2_T200_betaT0.02_uncond/checkpoint/800000.pkl}}\textsuperscript{,}\footnote{\url{https://drive.google.com/drive/folders/110CA22rwu_3EcllPYGhql0TnYeOBY77d} (\texttt{ema\_0.9999\_200000.pt})}\textsuperscript{,}\footnote{\url{https://drive.google.com/drive/folders/1bwJ9_S8HW7oq1zNPW2lDLNUrKcW-_mRQ} (\texttt{chair\_1799.pth})}. For each model, we generate 200 sampling trajectories using a deterministic 200-step Euler sampler with its default noise schedule, and transform them into variance-exploding (VE) coordinates, i.e., the coordinates of the drift-free diffusion. 
The terminal noise levels are $T=80$, $2.56$, $21.2$, and $157$ for EDM, DiffWave, Diffusion-LM, and PVD, respectively. The effective data dimension $r_{\text{eff}}\coloneq (\operatorname{tr}\bfLambda)^2/\operatorname{tr}(\bfLambda^2)$ is $9.2$ for CIFAR-10, $1136.5$ for SC09, $48.1$ for E2E, and $83.8$ for ShapeNet Chair. 


\begin{itemize}
    \item CIFAR-10 contains 50,000 RGB training images of resolution $32\times32$ in 10 classes, i.e., the data belong to $\bbR^{3\times 32\times 32}$. Following the EDM\footnote{\url{https://github.com/nvlabs/edm}} preprocessing protocol, we linearly rescale pixel values to $[-1,1]$ and estimate the mean and a rank-1,024 approximation of the covariance from the entire training set. 
    \item 
    SC09 contains the 38,908 utterances of the digits from ``zero'' to ``nine'', of which 31,158 utterances by 2,032 speakers form the training split~\citep{kong2021diffwave}. Following the released DiffWave--SaShiMi\footnote{\url{https://github.com/lmnt-com/diffwave}}\textsuperscript{,}\footnote{\url{https://github.com/albertfgu/diffwave-sashimi}} loader, we truncate or right-pad each utterance to 16,000 samples, i.e., the data belong to $\bbR^{1\times 16000}$. We estimate the mean and a rank-4,096 approximation of the covariance from all 38,908 utterances, since the released loader also trains on all of them. 
    \item E2E contains more than 50,000 crowdsourced restaurant descriptions, each paired with a meaning representation of 3 to 8 attributes, such as food type and price range. Following the Diffusion-LM\footnote{\url{https://github.com/XiangLi1999/Diffusion-LM}} preprocessing protocol, we use only the 42,061 descriptions in the training split, add \texttt{START} and \texttt{END} tokens to each description, and truncate or right-pad it to 64 tokens. Each token ID is mapped through the checkpoint's word-embedding matrix to a 16-dimensional embedding, i.e., the data belong to $\bbR^{64\times 16}$. We estimate the mean and the full covariance, of rank 1,024, and include the isotropic timestep-zero embedding noise used during training. During sampling, each predicted clean token embedding is mapped to its nearest vocabulary embedding before it enters the next update. 
    \item ShapeNetCore, a subset of the ShapeNet repository of 3D CAD models, contains about 51,300 models in 55 categories. We use its Chair point clouds, each with 15,000 points sampled uniformly from the mesh surface, and the official split with 4,612 training shapes. Following the PVD\footnote{\url{https://github.com/alexzhou907/PVD}} preprocessing protocol, we subtract the per-axis mean and divide by a single standard deviation, both computed over the training split, and keep the first 10,000 points of each shape for training. Each model input contains 2,048 points sampled uniformly with replacement from these 10,000 points, i.e., the data belong to $\bbR^{2048\times3}$. We average over this random subsampling in closed form when estimating the mean and the covariance, which has full rank 6,144. When each point cloud is flattened coordinate by coordinate, the covariance is $\bfW\otimes\bfI_{2048}+\bfB\otimes\mathbf{1}\mathbf{1}^{\top}$, where $\otimes$ is the Kronecker product, $\mathbf{1}\in\bbR^{2048}$ is the all-ones vector, $\bfW\in\bbR^{3\times 3}$ is the mean within-shape covariance, and $\bfB\in\bbR^{3\times 3}$ is the covariance of the shape centroids. It has three common-point modes, in which all points move together, with the eigenvalues of $\bfW+2048\,\bfB$, and three residual eigenspaces of dimension 2,047, in which the points move relative to their centroid, with the eigenvalues of $\bfW$. The predicted clean point cloud is not clipped.
\end{itemize}

For CIFAR-10 and SC09, the covariance is estimated by randomized principal component analysis with the stated rank. Directions outside the retained rank are given zero data variance, so the Gaussian flow contracts the initial noise along them toward the data mean in proportion to $t$. The isotropic Gaussian flow instead uses the trace of the full sample covariance.

\subsubsection{Trajectory visualization} 
Sampling trajectories are hard to compare and describe precisely in the original space. Beyond the subspace projection of our previous studies~\citep{chen2024trajectory,chen2025geometric}, we introduce the Gaussian reference frame as a new visualization technique.
\begin{itemize}
    \item \textbf{Subspace projection}. For each modality, we randomly select 32 of the 200 trajectories. Each real trajectory is translated to a common starting point and projected onto its own three leading temporal principal components, obtained by principal component analysis (PCA). Since independently estimated PCA bases have arbitrary orientations, we choose one real trajectory as the reference and align the others to it using the standard orthogonal Procrustes transformation fitted over the complete trajectory. Fitting this transformation in the three-dimensional PCA space or in the original space gives similar results. 
    \item \textbf{Two-eigenmode projection}. Since the standardized Gaussian trajectory has an invariant hyperbolic structure irrespective of the initial noise, we align each real trajectory with its Gaussian counterpart, which starts from the same noise, through one ambient-space orthogonal Procrustes transformation fitted over the complete trajectory, with the common initial noise subtracted before the fit and added back afterwards. We then project the aligned trajectory onto two eigenvectors $\bfu_i$ and $\bfu_j$ of the data covariance and divide each coordinate by its initial stochasticity $\tilde{y}_{T,i}$ or $\tilde{y}_{T,j}$, so that the Gaussian flow (dashed) traces the hyperbola of Eq.~\ref{eq:hyperbola}, from near $(T,T)$ at the initial noise to $(\sqrt{\lambda_i},\sqrt{\lambda_j})$ at the data end. The residual of this alignment depends only on the Gram kernels of the two start-aligned trajectories, so a trajectory with the same Gram kernel but unrelated directions would follow the hyperbolas exactly. The projection therefore visualizes, mode by mode, the shape mismatch that the ``After'' columns of Figure~\ref{fig:hyperbolic} quantify, whereas the ``Before'' columns compare the unaligned trajectories. The Gaussian trajectory, however, spans fewer directions than the real one: at most six for PVD, whose covariance has six distinct eigenvalues, and effectively few for the other models, whose eigenmodes with similar eigenvalues have nearly identical time profiles. The alignment thus leaves the rest of the real trajectory free to rotate without changing the residual, so the curves show one of several equally optimal views; the ``After'' columns are the same for all of them. We select four eigenmode pairs. The reference mode $j$ is the 1,000th eigenvector for CIFAR-10 and E2E and the 4,000th for SC09, and the mode $i$ is chosen such that $\Delta\lambda$ is approximately 1, 0.5--0.6, 0.3, and 0.1 times the gap of the second eigenmode; the leading eigenmode is excluded because its gap is far larger than all others. These four curves sample a nested family of hyperbolas indexed by $\Delta\lambda$: all start near $(T,T)$, share the diagonal as their asymptote, and end at height $\sqrt{\lambda_j}$; a larger gap bends the curve farther from the diagonal. Every other choice of $i$ yields a hyperbola of this family, lying between two plotted ones when its gap is between theirs, and $\lambda_i=\lambda_j$ gives the diagonal itself. For PVD, we compare the second and third common-point modes and the first two residual eigenspaces with the third residual eigenspace serving as the reference mode $j$. Solid lines show the pointwise mean over the real trajectories whose initial stochasticity satisfies $|\tilde{y}_{T,k}|\ge 0.01$ for all plotted modes (187 to 193 of 200), and thin lines show the ten real trajectories whose deviation from this mean is closest to the median.
\end{itemize} 

\subsubsection{Trajectory deviations} 
A dominant first principal component and trajectory congruence alone cannot validate a theoretical model, since simple baselines share both properties in high dimensions. We therefore compare real trajectories with their Gaussian counterparts and with three simpler references, all starting from the same initial noise.
Let $\Delta\bfx_t=\bfx_t-\bfx_T$ and $\Delta\barx_t=\barx_t-\barx_T$ denote the start-aligned real and Gaussian paths. Define progress as $u_t=(T-t)/(T-t_{\min})$, where $t_{\min}$ is the noise level of the final saved state. The Gaussian chord is
\begin{equation}
    \bfc_t=u_t\Delta\barx_{t_{\min}},
\end{equation}
so it uses the same start, endpoint, and noise-level progress as the Gaussian counterpart while discarding its curved shape. We also include a Brownian bridge around this Gaussian chord as the null shape model, to control for potential artifacts of principal-component-based dimensionality reduction,
\begin{equation}
    \bfb_t=\bfc_t+\sigma_B(\bfw_{u_t}-u_t\bfw_1),
\end{equation}
where $\bfw_u$ is a standard Wiener process in $\bbR^d$, so the bridge is pinned to the same two endpoints. Its amplitude is chosen independently for each real trajectory according to
\begin{equation}
    \sigma_B^2=
    \frac{\bbE_t\|\Delta\bfx_t-\bfc_t\|_2^2}
    {d\,\bbE_t\left[u_t(1-u_t)\right]},
\end{equation}
where $\bbE_t$ is the uniform average over the saved states, i.e., $\nu$ is their empirical distribution. Thus, the bridge matches the real trajectory's mean squared deviation from the Gaussian chord but retains a random shape. Since the random shape $\bfw_{u_t}-u_t\bfw_1$ is independent of the real trajectory, which sets only its amplitude $\sigma_B$, the bridge's mean squared deviation from the real trajectory before alignment is about twice that of the chord, giving the ratio of about $\sqrt{2}$ in Figure~\ref{fig:hyperbolic}. The isotropic Gaussian flow replaces $\bfSigma$ with the scaled identity $(\operatorname{tr}\bfSigma/d)\,\bfI_d$, which keeps the total data variance and the noise-level dependence of the Gaussian flow but none of its covariance structure; it moves along a straight line from the initial noise toward the data mean.

\begin{figure*}[t]
    \centering
    \subfloat[Image. 
    \label{fig:hist_all_cifar10}]{%
        \begin{minipage}[b]{0.235\textwidth}%
            \includegraphics[width=\linewidth]{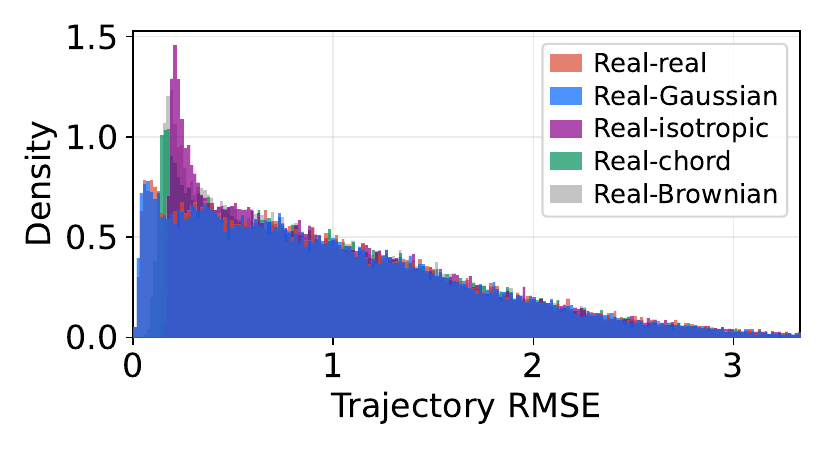}\\[-4pt]%
            \includegraphics[width=\linewidth]{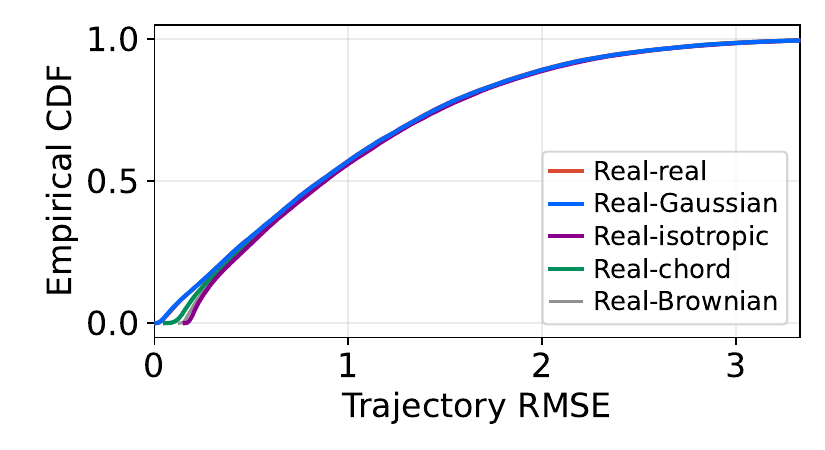}%
        \end{minipage}%
    }%
    \hfill
    \subfloat[Audio. 
    \label{fig:hist_all_diffwave}]{%
        \begin{minipage}[b]{0.235\textwidth}%
            \includegraphics[width=\linewidth]{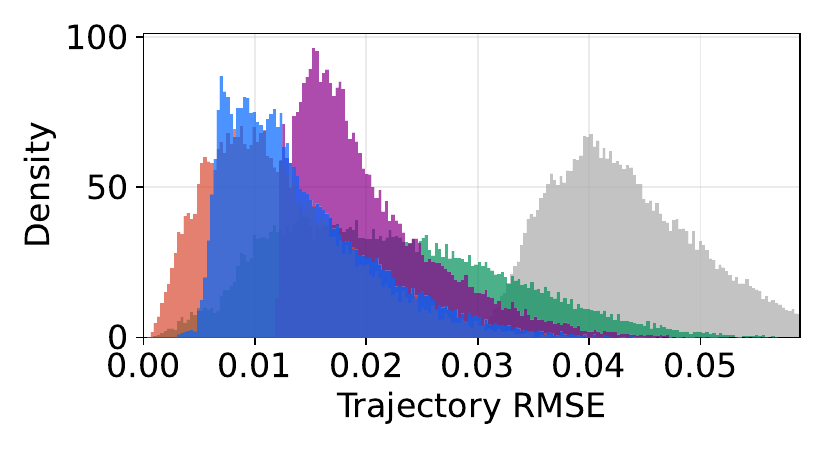}\\[-4pt]%
            \includegraphics[width=\linewidth]{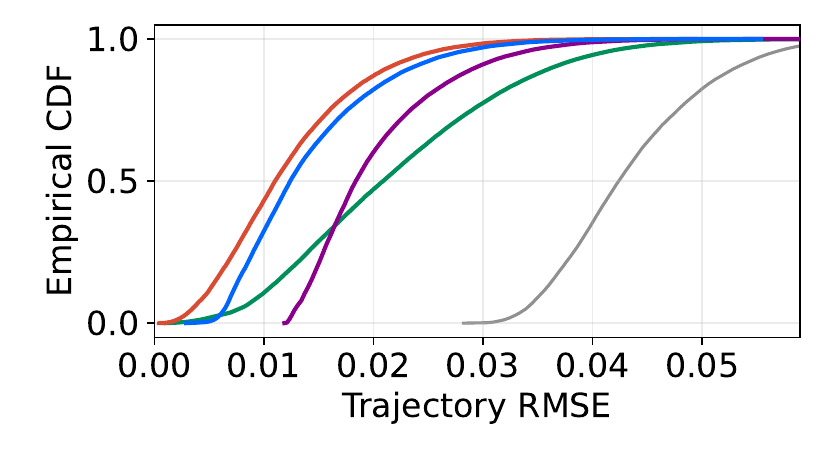}%
        \end{minipage}%
    }%
    \hfill
    \subfloat[Text. 
    \label{fig:hist_all_diffusion_lm}]{%
        \begin{minipage}[b]{0.235\textwidth}%
            \includegraphics[width=\linewidth]{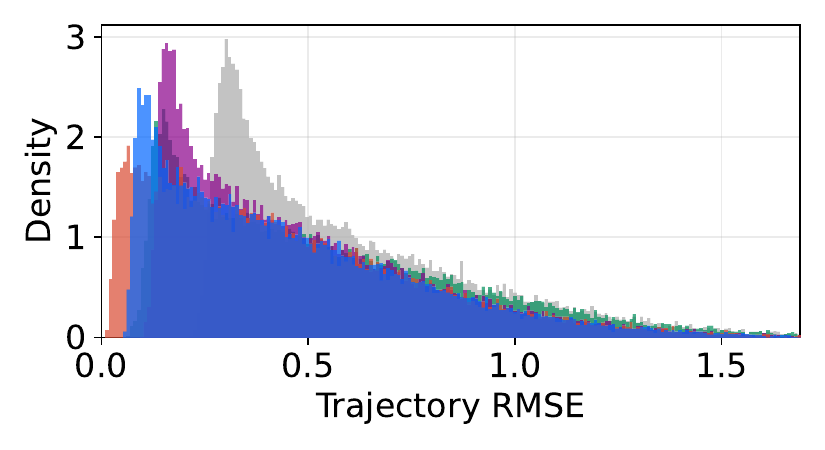}\\[-4pt]%
            \includegraphics[width=\linewidth]{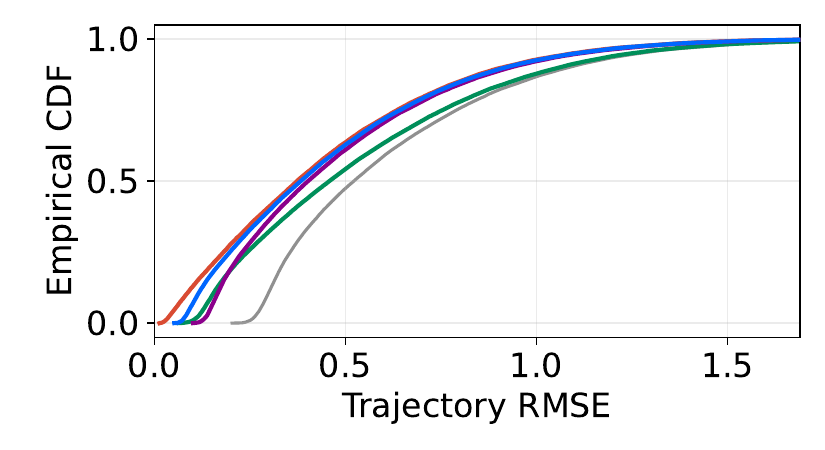}%
        \end{minipage}%
    }%
    \hfill
    \subfloat[3D. 
    \label{fig:hist_all_pvd}]{%
        \begin{minipage}[b]{0.235\textwidth}%
            \includegraphics[width=\linewidth]{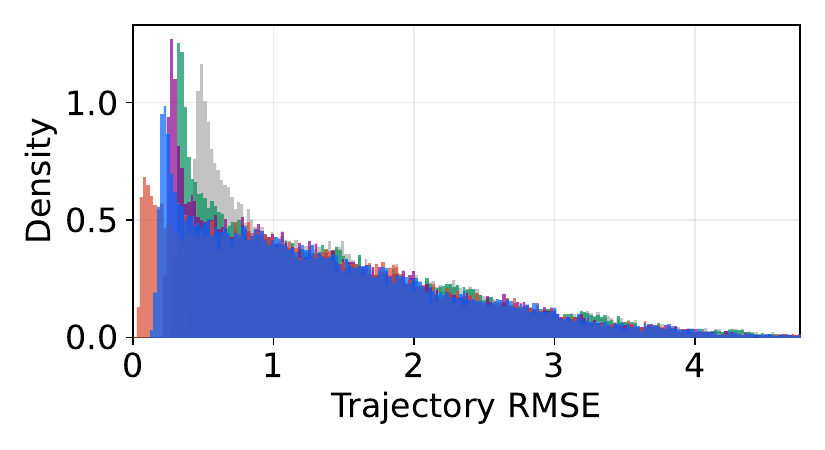}\\[-4pt]%
            \includegraphics[width=\linewidth]{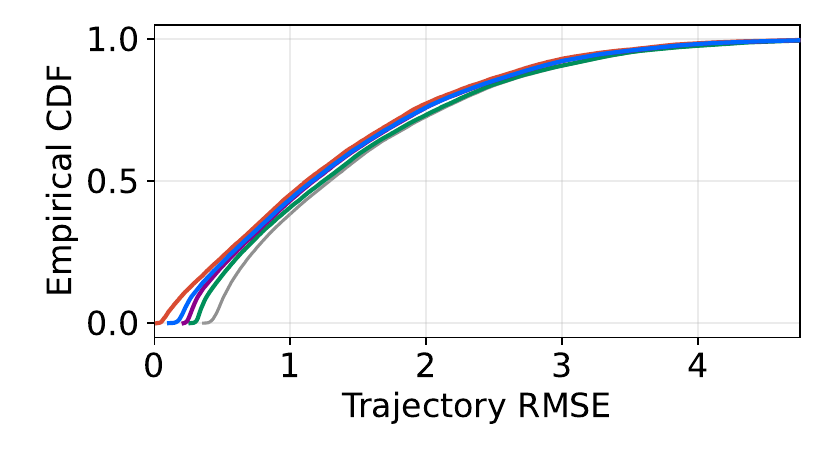}%
        \end{minipage}%
    }%
    \caption{Distributions of the trajectory RMSE for independently initialized paths. The top row shows the error densities, and the bottom row shows the corresponding empirical cumulative distribution functions. In addition to the real--real and real--Gaussian comparisons in the bottom row of Figure~\ref{fig:hyperbolic}, we include the isotropic Gaussian flow, Gaussian chord, and Brownian bridge.}
    \label{fig:hist_cdf_all}
\end{figure*}

\begin{figure*}[t]
    \centering
    \subfloat[Image. 
    \label{fig:cdf_s_cifar10}]{%
        \includegraphics[width=0.235\textwidth]{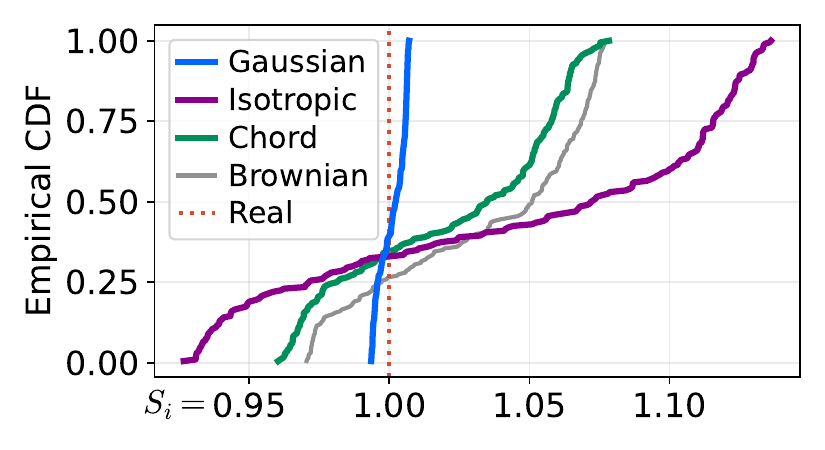}%
    }%
    \hfill
    \subfloat[Audio. 
    \label{fig:cdf_s_diffwave}]{%
        \includegraphics[width=0.235\textwidth]{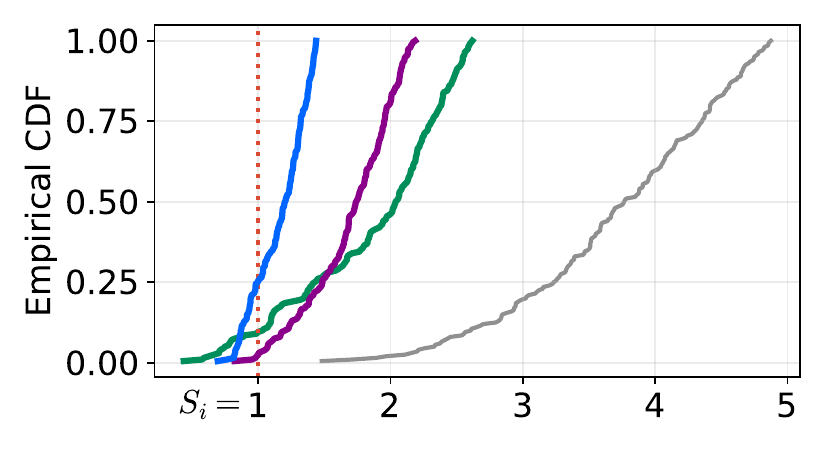}%
    }%
    \hfill
    \subfloat[Text. 
    \label{fig:cdf_s_diffusion_lm}]{%
        \includegraphics[width=0.235\textwidth]{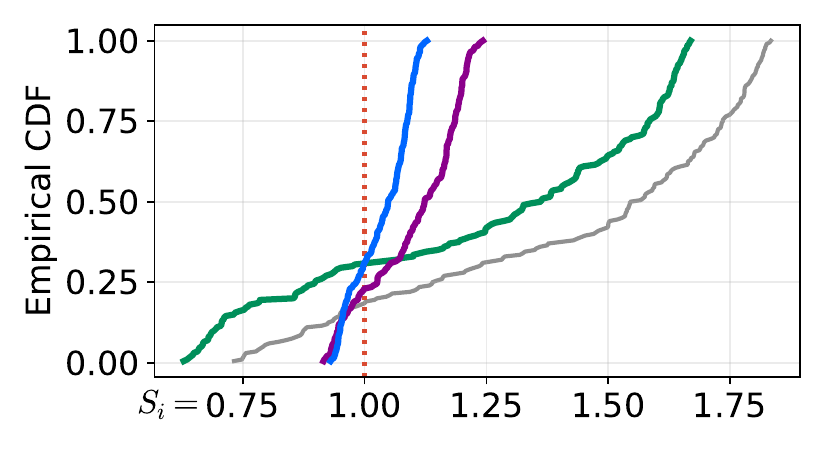}%
    }%
    \hfill
    \subfloat[3D. 
    \label{fig:cdf_s_pvd}]{%
        \includegraphics[width=0.235\textwidth]{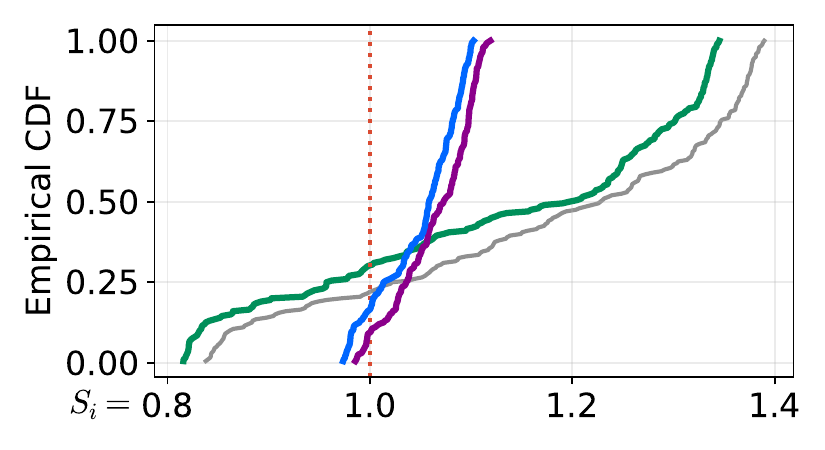}%
    }%
    \caption{For a reference family $\mathcal M$, we define
        $
        S_i^{(\mathcal M)}
        =
        \operatorname{mean}_{j\neq i}
        d(\mathbf X_i,\mathbf M_j) / 
        \operatorname{mean}_{j\neq i}
        d(\mathbf X_i,\mathbf X_j),
        $
    where $d(\cdot, \cdot)$ is the optimally aligned trajectory RMSE and $\mathbf M_j$ uses the same initialization as real trajectory $\mathbf X_j$. The curves show the distribution of this score over the 200 real trajectories. The dotted line marks $S_i=1$, matching the average real--real error; lower values mean closer agreement.}
    \label{fig:cdf_s}
\end{figure*}
In both comparisons below, every trajectory is translated so that its initial state lies at the origin, as for $\Delta\bfx_t$ above, and its $L$ saved states form the rows of a trajectory matrix in $\mathbb R^{L\times d}$.
\begin{itemize}
    \item \textbf{Identical-start trajectory deviation}.
We evaluate the Gaussian counterpart, the isotropic Gaussian flow, the Gaussian chord, and the Brownian bridge before and after alignment. For any two trajectory matrices $\mathbf A,\mathbf B\in\mathbb R^{L\times d}$, we define the trajectory root-mean-square error (RMSE) over the saved states as
\begin{equation}
    d(\mathbf A,\mathbf B)=\min_{\bfQ^\top\bfQ=\bfI}
    \left[\frac{1}{Ld} \left\|
    \mathbf A-\mathbf B\bfQ
    \right\|_{\mathrm F}^2
    \right]^{1/2},
    \label{eq:independent-start-error}
\end{equation}
where the minimum over $\bfQ$ is the alignment, a single ambient-space orthogonal Procrustes transformation fitted over the complete trajectory, and $\bfQ=\bfI_d$ before alignment. We normalize each deviation by that of the Gaussian chord before alignment, so the chord before alignment has ratio one by construction, and a ratio below one indicates an improvement over this endpoint-matched linear reference. The table in Figure~\ref{fig:hyperbolic} reports the mean of these ratios over the 200 trajectories of each model.
    \item \textbf{Independent-start trajectory deviation}. In addition, we compare independently initialized paths. This setting is the more realistic one, since real trajectories from different initial noise are themselves congruent only up to an orthogonal transformation (Figure~\ref{fig:vis}(c) and our previous study~\citep{chen2024trajectory}). Let $\mathbf X_i$ denote the trajectory matrix of real trajectory $i$, and let $\mathbf G_j$ denote that of the Gaussian trajectory starting from the same noise as real trajectory $j$. The bottom row of Figure~\ref{fig:hyperbolic} reports the distributions of $d(\mathbf X_i,\mathbf X_j)$ and $d(\mathbf X_i,\mathbf G_j)$ over all $200\times199$ ordered pairs with $i\neq j$, and Figure~\ref{fig:hist_cdf_all} adds the other references. In Figure~\ref{fig:cdf_s}, the scores of the Gaussian counterpart are the most concentrated and lie closest to one, matched only by the isotropic Gaussian flow in (d). 
\end{itemize}

Together, the identical-start and independent-start comparisons support the hyperbolic structure of the Gaussian flow as a reasonable approximation of real trajectories.

As discussed in the main text, at $t=s$ the relative fluctuation of Gaussian sampling trajectories (Eq.~\ref{eq:fluctuation}) equals $\sqrt{2/r_{\text{eff}}}$ at the data end and decays quickly as $t$ grows, since adding $t^2$ to every eigenvalue flattens the spectrum. Once $t^2$ exceeds the root-mean-square eigenvalue $\sqrt{\operatorname{tr}(\bfLambda^2)/d}$, it reaches the $\mathcal{O}(d^{-1/2})$ level, i.e., $\sqrt{2/d}\le\sqrt{\operatorname{Var}(G(t,t))}/\bbE[G(t,t)]\le 2/\sqrt{d}$: the lower bound follows from the Cauchy--Schwarz inequality, and the upper bound from $\operatorname{tr}(\bfLambda^2)\le d\,t^4$. At the data end, we note the following for power-law eigenspectra.
\begin{remark}
    The eigenspectrum of natural images typically exhibits a power-law decay, $\lambda_i \propto i^{-\alpha}$. In the slow-decay regime ($0\leq \alpha<1$), the quantity $r_{\text{eff}}^{-1/2}$ converges to zero as $\calO(d^{-1/2})$ for $\alpha<1/2$, as $\calO(\sqrt{\ln d/d})$ at $\alpha=1/2$, and as $\calO(d^{-(1-\alpha)})$ for $1/2<\alpha<1$. At the critical boundary ($\alpha=1$), it converges to zero as $\calO(1/\ln d)$. In the steep-decay regime ($\alpha>1$), it converges to a finite constant given by the Riemann zeta function $\zeta(\cdot)$: $\lim_{d\to \infty} r_{\text{eff}}^{-1/2}= \sqrt{\zeta(2\alpha)}/\zeta(\alpha)$, which monotonically approaches zero as $\alpha\to 1^+$.
\end{remark}

\end{appendices}

\end{document}